\documentclass{article}

\usepackage{microtype}
\usepackage{booktabs} 

\usepackage{hyperref}

\usepackage[accepted]{icml2019}

\icmltitlerunning{Enhancing Gradient-based Attacks with Symbolic Intervals}

\usepackage{comment}
\usepackage{multirow}
\usepackage{setspace}
\usepackage{graphicx}
\RequirePackage{subcaption}
\usepackage{amsmath}
\usepackage{amssymb}

\usepackage[all=normal, title=normal, paragraphs=tight, wordspacing=normal]{savetrees}

\begin{document}

\twocolumn[
\icmltitle{Enhancing Gradient-based Attacks with Symbolic Intervals}
 \vspace{-10pt}
 
 \begin{icmlauthorlist}
\icmlauthor{Shiqi Wang}{ed}
\icmlauthor{Yizheng Chen}{ed}
\icmlauthor{Ahmed Abdou}{to}
\icmlauthor{Suman Jana}{ed}
\end{icmlauthorlist}

 \vspace{-10pt}
 
\icmlaffiliation{ed}{Columbia University}
\icmlaffiliation{to}{Pennsylvania State University}

\icmlcorrespondingauthor{Shiqi Wang}{tcwangshiqi@cs.columbia.edu}

\icmlkeywords{Machine Learning, ICML}

\vskip 0.3in
]
\printAffiliationsAndNotice{}




\begin{abstract}
Recent breakthroughs in defenses against adversarial examples, like adversarial training, make the neural networks robust against various classes of attackers (e.g., first-order gradient-based attacks). However, it is an open question whether the adversarially trained networks are truly robust under unknown attacks. In this paper, we present interval attacks, a new technique to find adversarial examples to evaluate the robustness of neural networks. Interval attacks leverage symbolic interval propagation, a bound propagation technique that can exploit a broader view around the current input to locate promising areas containing adversarial instances, which in turn can be searched with existing gradient-guided attacks.
We can obtain such a broader view using sound bound propagation methods to track and over-approximate the errors of the network within given input ranges.
Our results show that, on state-of-the-art adversarially trained networks, interval attack can find on average 47\% relatively more violations than the state-of-the-art gradient-guided PGD attack.

\end{abstract}

\vspace{-20pt}
\section{Introduction}
\label{sec:intro}

Deep learning systems have achieved strong performance at large scale. However, attackers can easily locate adversarial examples that are perceptibly the same as the original images and misclassified by state-of-the-art deep learning models. Making Neural Networks (NNs) robust against adversarial inputs has resulted in an arms race between new defenses and attacks that break them. 
Recent breakthroughs in defenses can make NNs robust against various classes of attackers, including state-of-the-art \texttt{Projected Gradient Descent} (PGD) attack. In~\cite{madry2017towards}, Madry et al. argued that PGD attacks are the ``ultimate'' first-order attacks by preliminary tests. They adversarially retrain the networks with violations found by PGD to obtain robustness. Such training procedure is called adversarial training. The conclusion is that training models with PGD attacks can gain robustness against all first-order gradient attacks.

Nonetheless, the adversarially trained networks might not be truly robust against unknown attacks even if it has good robustness against state-of-the-art attacks like PGD. Since neural networks are highly non-convex, there is high chance that the existing attacks will miss many adversarial examples. 

The main challenge behind locating adversarial examples using gradients as guidance is that the search process can get stuck at local optima. Therefore, we hope to have a special first-order gradient that can offer a broader view within surrounding area, potentially guiding us towards the worst-case behavior of neural networks.

Our critical insight is that we can obtain such a broader view by existing sound over-approximation methods. These methods are sound. It means the estimated output ranges are guaranteed to always over-approximate the ground-truth ranges (i.e., never miss any adversarial examples but might introduce false positives).
For convenience, we call these methods \texttt{sound bound propagation}.
Recently, they have been successfully used for verifiable robust training~\cite{wong2018scaling,mirman2018differentiable,wang2018mixtrain,gowal2018effectiveness}. For each step of training, the weights of the networks are updated with the gradients of the verifiable robust loss provided by sound propagation methods over given input ranges.
We borrow a similar idea for our attacks by proposing a generic way to extract and apply the gradient from the output provided by these methods.

\begin{figure}[!hbt]
	\centering
	\begin{subfigure}[t]{0.48\columnwidth}
		\centering
		\includegraphics[width=\columnwidth]{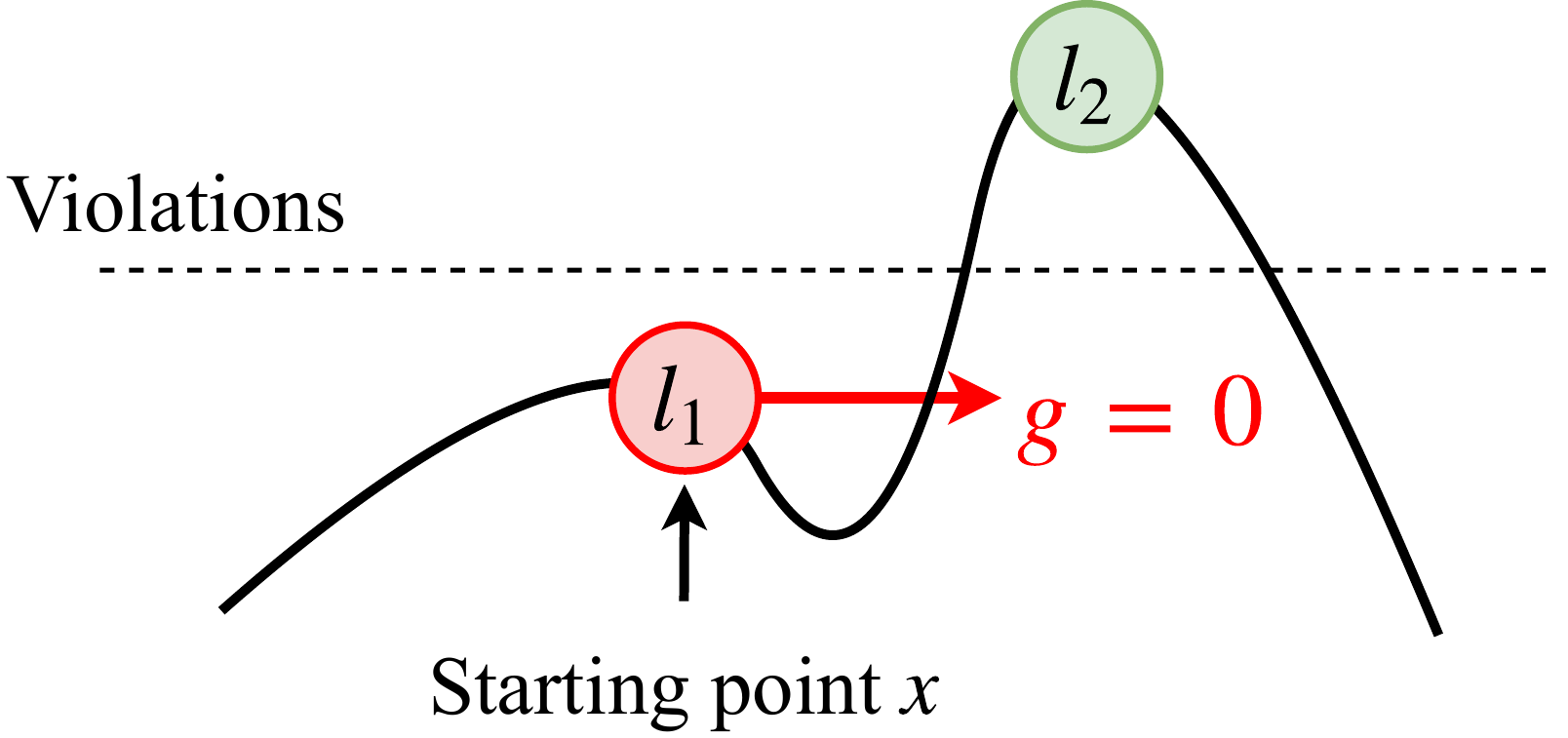}
		\caption{PGD attack}
		\label{subfig:iapgd}
	\end{subfigure}
	\centering
	\begin{subfigure}[t]{0.48\columnwidth}
		\centering
		\includegraphics[width=\columnwidth]{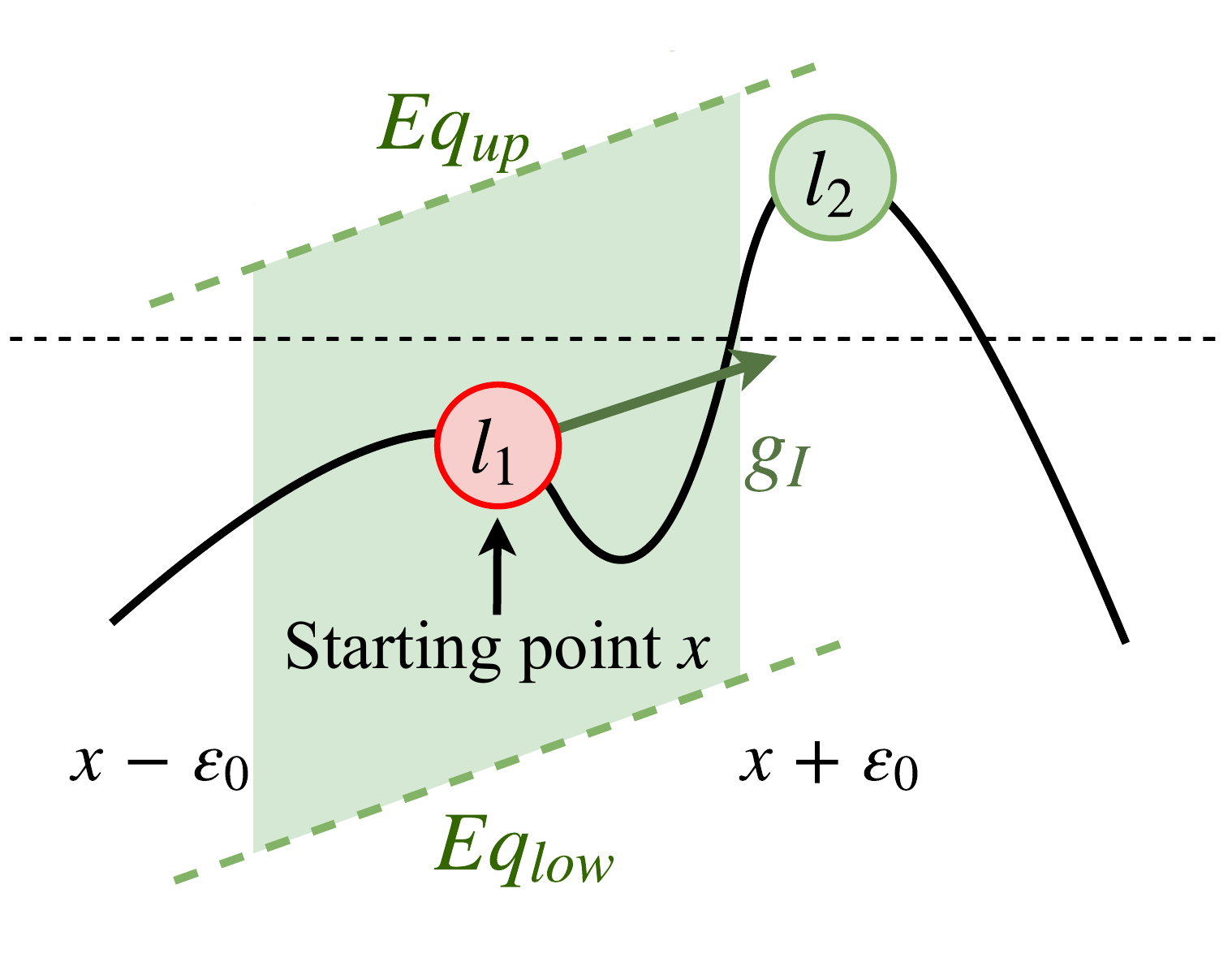}
		\caption{Interval attack}
		\label{subfig:ia}
	\end{subfigure}
	\caption{\bf \small The difference between regular gradient $g$ and interval gradient $g_I$. Figure (a) illustrates that PGD attacks using regular gradients might get stuck at $l_1$. Figure (b) shows that interval gradient $g_I$ over an input region $B_{\epsilon_0}$ allows us to avoid such problem and successfully locate the violation $l_2$. Here $Eq_{up}$ and $Eq_{low}$ are the symbolic upper and lower bounds of the output found by one type of sound bound propagation method, symbolic interval analysis~\cite{reluval2018,shiqi2018efficient}.}
	\label{fig:ia}
	\vspace{-20pt}
\end{figure}

Essentially, sound propagation methods relax the input range into various abstract domains (e.g., zonotope~\cite{gehrai}, convex polytope~\cite{wong2018provable}, and symbolic interval~\cite{reluval2018,shiqi2018efficient}) and propagate them layer by layer until we have an over-approximated output abstract domain. The gradient that encodes a broader view within surrounding area can be obtained from such abstract domains of the outputs. It can guide us toward a promising sub-area. In Figure~\ref{fig:ia}, we show the symbolic interval and its interval gradient $g_I$ as an example to illustrate their effectiveness on the attack performance.

\begin{figure*}[!hbt]
	\centering
	\begin{subfigure}[t]{0.9\columnwidth}
		\includegraphics[width=\columnwidth]{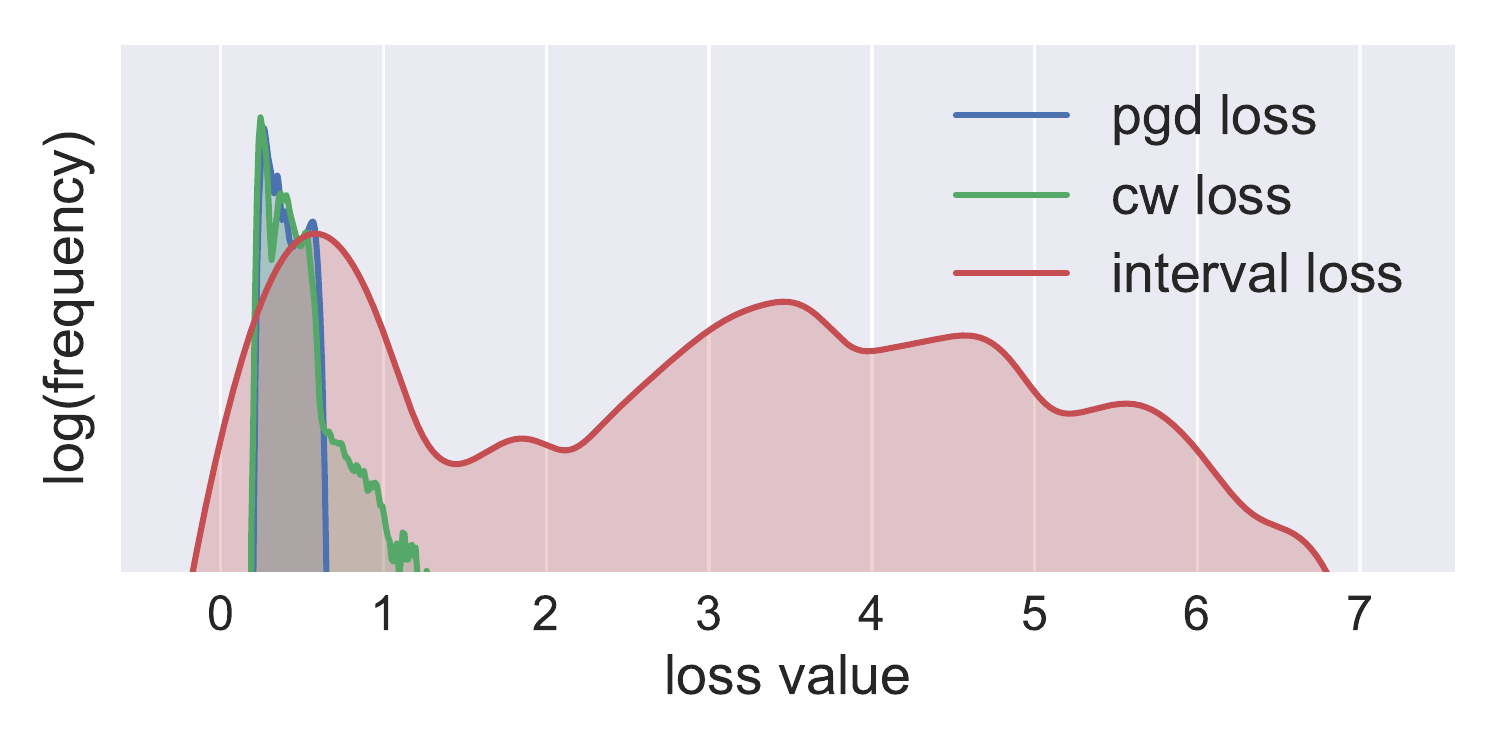}
		\vspace{-20pt}
		\caption{\bf \small MNIST image 12}
		\label{fig:iaimg1}
	\end{subfigure}
	\begin{subfigure}[t]{0.9\columnwidth}
		\includegraphics[width=\columnwidth]{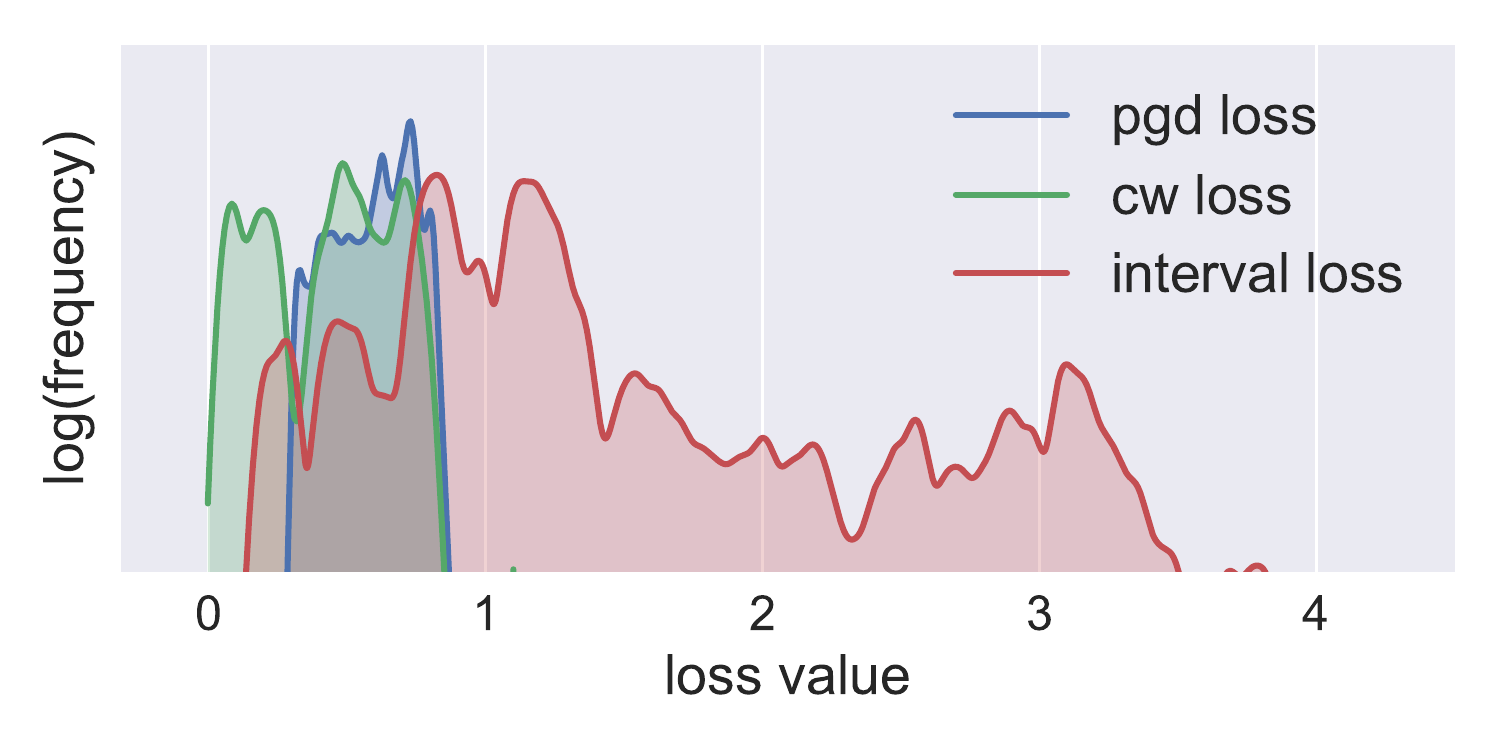}
		\vspace{-20pt}
		\caption{\bf \small MNIST image 31}
		\label{fig:iaimg2}
	\end{subfigure}
\vspace{-7pt}
	\caption{\bf \small Distributions of the loss values from robustness violations found by PGD attacks (blue), CW attacks (green), and interval attack (red) with 100,000 random starts within the allowable input range $B_\epsilon(x)$. The loss values found by CW and PGD attacks are very small and concentrated. However, interval attack shows there are still many distinct violations with much larger loss values.}
	\label{fig:iaimg}
	\vspace{-15pt}
\end{figure*}


In this paper, we describe such a novel gradient-based attack, \texttt{interval attack}, which is the first attack framework to combine both sound bound propagation method and existing gradient-based attacks. Here, we focus specifically on symbolic interval analysis as our sound bound propagation. However, our attack is generic. It can be adapted to leverage any other sound bound propagation methods and their corresponding gradient values. 
The interval attack contains two main steps: it (1) uses interval gradient to locate promising starting points where the surrounding area has the worst-case behaviors indicated by the sound propagation methods, and (2) uses strong gradient-based attacks to accurately locate the optima within the surrounding area.
Such design is due to the fact that the sound bound propagation will introduce overestimation error and prevent convergence to an exact optima if only relying on interval gradient. Specifically, such overestimation error is proportional to the width of the input range (e.g., $\epsilon_0$ in Figure~\ref{fig:iaimg}).
To achieve the best performance by encoding the broader view of the surrounding area, we design a strategy (Section~\ref{sec:ia}) to dynamically adjust the range used for each step of the interval attack.

We implement our attack with symbolic interval analysis and evaluate it on three MNIST models that are trained with the state-of-the-art defense, adversarial training~\cite{madry2017towards}. Interval attack is able to locate on average 47\% relatively more adversarial examples than PGD attacks. We also evaluate our attack on the model from MadryLab MNIST Challenge~\cite{mnistchallenge}, which is considered to be the most robust model on the MNIST dataset so far. On this model, interval attack has achieved the best result than all of the existing attacks. Code of interval attack is available at: \url{https://github.com/tcwangshiqi-columbia/Interval-Attack}.

\section{Interval attack}
\label{sec:ia}


In this section, we first motivate the need for stronger attacks by analyzing the loss distribution found by PGD attacks with $10^5$ different random starting points. Then we dive into the definition of interval gradients and details of interval attack procedure.

\noindent{\bf Why we need stronger attack?} PGD attack is so far a well-known and a standard method to measure the robustness of trained NNs. Madry et al. even argued that PGD attacks are the ``ultimate'' first-order attacks, i.e., no other first-order attack will be able to significantly improve over the solutions found by PGD~\cite{madry2017towards}. In order to support the statement, Madry et al. performed $10^5$ iterations of the PGD attack from random starting points within bounded $L_{\infty}$-balls of test inputs and showed that all the solutions found by these instances are distinct local optima with similar loss values. Therefore, they concluded that these distinct local optima found by PGD attacks are very close to the best solution that can be found by any first-order attacker.

In Figure~\ref{fig:iaimg}, we demonstrate such assumption to be flawed with our first-order interval attacks. Essentially, for each random starting point, instead of directly applying PGD attack, interval attack will first use sound bound propagation to locate promising sub-area and then apply PGD attacks. We randomly picked two different images on which PGD attacks cannot find violations but the interval attacks can, and then we repeated the same tests in~\cite{madry2017towards} for PGD, CW, and interval attacks. On both images, regular PGD attacks cannot find any violation even with $10^5$ random starts. The losses found by PGD attacks are very concentrated between 0 and 1, which is consistent with the observations in~\cite{madry2017towards}. Similarly, the losses of CW attacks on these images show concentrated distribution.
However, interval attacks can locate promising sub-area and use the same PGD attacks to offer a much larger range of loss distribution. Particularly, it can find over $50,000$ violations out of the same $10^5$ starting points.
Therefore, the method of Madry et al.~\cite{madry2017towards} cannot provide robustness against \emph{all} first-order adversaries even if the model is robust against PGD attacks with $10^5$ random starting points.

\noindent\textbf{Interval gradient ($g_I$).} As described in Section~\ref{sec:intro}, any existing sound bound propagation method can be a good fit in our interval attack. In this paper, we use {symbolic interval analysis} to provide tight output bounds~\cite{reluval2018,shiqi2018efficient}. Essentially, such analysis produces two parallel linear equations ($Eq_{up}$ and $Eq_{low}$ shown in Figure~\ref{fig:ia}) to tightly bound the output of each neuron. Compared to other types of sound bound propagation, symbolic interval analysis provides tight over-approximation while its interval gradient can be easily accessed.

Based on that, we define the interval gradient $g_I$ to be equal to the slope of these parallel lines. For example, let us assume symbolic interval analysis propagates input range $x=[0,1]$ and finally provides the bounds of a neural network's output as $[2x,2x+3]$ through the network. Then, the interval gradient will be  $\frac{d([2x,2x+3])}{dx}=2$. 

For other sound bound propagation methods that do not have parallel output bounds, we can estimate the interval gradients with average gradients (slope) of all the corresponding output bounds. Here we give one generic definition of interval gradients. Assume the output range provided by sound bound propagation is presented in the form of abstract domain as $Eq_i (i=1,...,n)$, then the interval gradient can be estimated by:
\vspace{-10pt}
$$g_I = \frac{1}{n}\sum_{i}^{n}\frac{d Eq_i}{d x}$$
\vspace{-20pt}

Note that how to choose the best sound bound propagation methods and how to estimate interval gradients worth further discussion. Here we just provide one possible solution that has good empirical results with symbolic interval analysis.

\begin{algorithm}[!hbt]
	\caption{Interval Attack}
	\label{alg:ia}
	
	\begin{tabular}{|l|}
		\hline
		\textbf{Inputs}: $x$ $\leftarrow$ target input, $\epsilon$ $\leftarrow$ attack budget \ \\
		\textbf{Parameters}: \textbf{$t$}: iterations, \textbf{$\alpha$}: step size, \textbf{$\epsilon_0$}: starting point, \\ {p}: input region step size \\
		\textbf{Output}: $x'$ $\leftarrow$ perturbed $x$ \ \\
		\hline
	\end{tabular}
	\small
	\begin{spacing}{0.9}
		\begin{algorithmic}[1]
			\STATE $x' = x$
			\FOR {$i \gets 1$ to $t$}
			\STATE $\epsilon_0'=\epsilon_0$
			\STATE // The smallest $\epsilon_0$ that can potentially cause violations
			\WHILE {Safe analyzed by sound bound propagation}
			\STATE $\epsilon_0'=\epsilon_0'*p$
			\IF {$\epsilon_0'>=\epsilon/2$}  
			\STATE break // Prevent $\epsilon_0$ being too large
			\ENDIF
			\ENDWHILE
			\STATE $g_I=\nabla_xEq$ // Interval gradient of symbolic intervals
			\STATE $x' = x'+\alpha g_I$
			\STATE // Bound $x'$ within allowable input ranges
			\STATE $x' = clip(x', x-\epsilon, x+\epsilon)$   
			\IF {$argmax(f_\theta(x'))\neq y$} 
			\STATE {\bf return} $x'$ 
			\ENDIF
			\ENDFOR
			\STATE PGD Attack($x'$) \label{alg:pgd_in_ia}
			\STATE {\bf return} $x'$
		\end{algorithmic}
	\end{spacing}
\end{algorithm}

\noindent\textbf{Interval attack details.}
Algorithm~\ref{alg:ia} shows how we implement the interval attack. The key challenge is to find a suitable value of $\epsilon_0$ representing the input region over which the interval gradient will be computed (Line 3 to Line 11)  at each iteration of the interval-gradient-based search.
If $\epsilon_0$ is too large, the symbolic interval analysis will introduce large overestimation error leading to the wrong
direction. On the other hand, if $\epsilon_0$ is too small, the information from the
surrounding area might not be enough to make a clever update.

Therefore, we dynamically adjust the size of $\epsilon_0$ during the attack. Specifically, we allow a hyperparameter $p$ to be tuned during the attack procedure which controls the step size for searching $\epsilon_0$ as Line 7. Since $\epsilon_0$ is usually small such that the sound bound propagation tends to be accurate. We can then pick the smallest $\epsilon_0$ that will cause potential violations in relaxed output abstract domain. The suitable interval gradient for that abstract domain can thus be accessed within such $\epsilon_0$.

After a bounded number of iterations with interval gradients, we use PGD attack with a starting point from the promising sub-area identified by the interval-gradient-based search to locate a concrete violation as shown in Line 21.

\section{Evaluation}
\label{sec:eval}

\begin{table*}[!hbt]
	\small
	\centering
	\begin{tabular}{|c|r|r|r|r|r|r|r|}
		\hline
		\multirow{2}{*}{Network} & \multirow{2}{*}{\# Hidden units} & \multirow{2}{*}{\# Parameters} & \multirow{2}{*}{ACC (\%)} & \multicolumn{4}{c|}{Attack success rate (\%)} \\ \cline{5-8} 
		&                               &                             &                               & PGD       & CW        & Interval Attack & Interval Attack Gain     \\ \hline
		MNIST\_FC1                    & 1,024                         & 668,672                     & 98.1        & 39.2      & 42.2     &\textbf{56.2}     &  {+17 (43\%)}   \\ \hline
		MNIST\_FC2                    & 10,240                        & 18,403,328               & 98.8        & 34.4      & 32.2   & \textbf{44.4}      &  {+10.0 (38\%)}       \\ \hline
		MNIST\_Conv                   & 38,656                        & 3,274,634               & 98.4        & 7.2      & 7.3  & \textbf{11.6}$^*$        &  {+4.4 (61\%)}     \\ \hline
		\multicolumn{8}{l}{\footnotesize * {Interval attack achieves the best attack success rate in MadryLab MNIST Challenge~\cite{mnistchallenge}}.}
	\end{tabular}
	\vspace{-5pt}
	\caption{\bf \small Attack success rates of PGD, CW and interval attack ($\epsilon=0.3$) on MNIST\_FC1, MNIST\_FC2 and MNIST\_Conv. All of the networks are adversarially robust trained using PGD attacks with $\epsilon=0.3$~\cite{madry2017towards}. The interval attack gain is computed as an increase over the success rate of PGD attack and the relative improvement percentage shown in parenthesis.}
	\label{tab:ia_mnist}
	\vspace*{-15pt}
\end{table*}


\noindent\textbf{Setup.} We implement symbolic interval analysis~\cite{shiqi2018efficient} and interval attack on top of \texttt{Tensorflow 1.9.0}
\footnote{https://www.tensorflow.org/}.
All of our experiments are run on a GeForce GTX 1080 Ti.

We evaluate the interval attack on three neural networks MNIST\_FC1, MNIST\_FC2 and MNIST\_Conv. The network details are shown in Table \ref{tab:ia_mnist}. 
All of these networks are trained to be adversarially robust using the Madry et al.'s technique~\cite{madry2017towards}. 
For PGD attacks, we use 40 iterations and 0.01 as step size. We define the robustness region, $B_\epsilon(x)$, to be bounded by $L_\infty$ norm with $\epsilon=0.3$ over normalized inputs (76 out of 255 pixel value). We use the 1,024 randomly selected images from the MNIST test set to measure accuracy and robustness. MNIST\_FC1 contains two hidden layers each with 512 hidden nodes and achieves 98.1\% test accuracy. Similarly, MNIST\_FC2 contains five hidden layers each with 2,048 hidden nodes and achieves 98.8\% test accuracy. MNIST\_Conv was adversarially robust trained by Madry et al.~\cite{madry2017towards} and was released publicly as part of the MadryLab MNIST Challenge~\cite{mnistchallenge}. The model uses two convolutional layers and two maxpooling layers to achieve 98.4\% test accuracy, on which the PGD attack success rate is only 7.2\%.



\noindent\textbf{Attack effectiveness.} The interval attack is very effective at finding violations in all of the MNIST models.
We compared the strength of the interval attack against the state-of-the-art PGD and CW attacks, as shown in Table~\ref{tab:ia_mnist}.
For the first two models, we give the same amount of attack time.
The iterations for PGD and CW are around 7,200 for MNIST\_FC1 and 42,000 for MNIST\_FC2.
We ran the interval attack for 20 iterations against MNIST\_FC1 and MNIST\_FC2,
which took 53.9 seconds and 466.5 seconds respectively. 
The interval attack is able to achieve from 38\% to 61\% relative increase in the attack success rate
of PGD.
On the MNIST\_Conv network from the MadryLab MNIST Challenge, interval attack achieves the highest success rate compared to all of the existing attacks.

Our results show that using only first-order gradient information,
the interval attack can significantly improve the attack success rate over that of the PGD and CW attacks in
adversarially robust networks.
\begin{figure}[!hbt!]
	\centering
	\includegraphics[width=0.6\columnwidth]{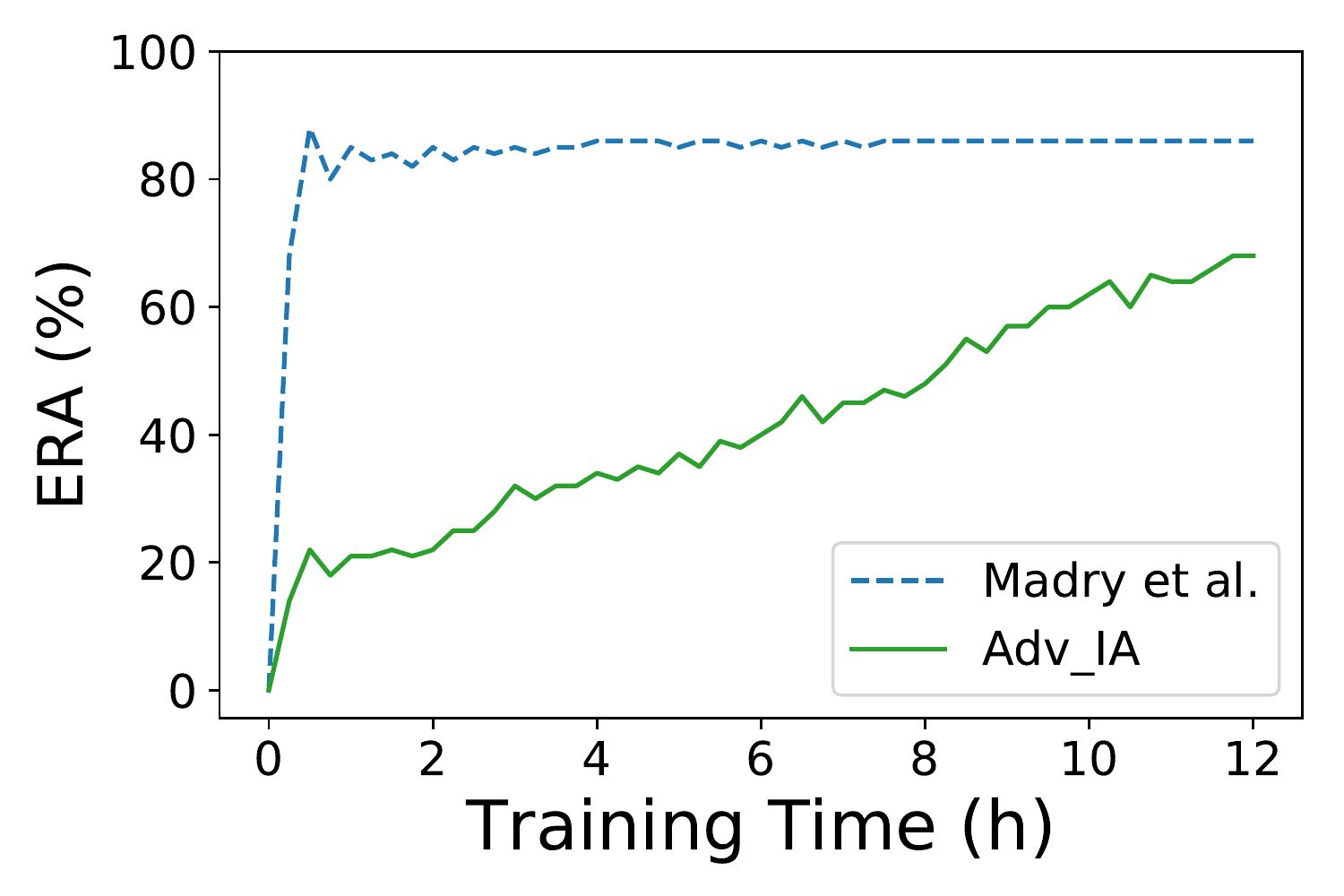}
	\vspace{-10pt}
	\caption{\bf \small Adversarially robust training using the interval attack does not converge well compared to using PGD attacks, given 12 hours of training time on the MNIST\_Small network with $L_\infty\leq 0.3$. The estimated robust accuracy of training using PGD attacks~\cite{madry2017towards} quickly converges to 89.3\% ERA, while training using the interval attack takes significantly longer to converge.}
	\label{fig:ia_adv}
	\vspace{-5pt}
\end{figure}

\noindent\textbf{Inefficiency of adversarially robust training with interval attack.}
One obvious way of increasing the robustness of trained networks against interval-based attacks is to use such attacks for adversarial training. 
However, due to the high overhead introduced by sound bound propagation methods, robust training with interval attack often struggle to converge. 
To demonstrate that, we evaluated adversarial robust training using interval attack on MNIST\_Small network for $L_\infty\leq 0.3$. As shown in Figure \ref{fig:ia_adv}, even after 12 hours of training time for such a small network, the interval-based adversarially robust training does not converge as well as its PGD-based counterpart. To achieve the same $80\%$ ERA, such interval-based adversarially robust training takes around 15.5 hours which is 47 times slower than its PGD-based counterpart. The gap will be further widened on larger networks.

Therefore, instead of improving the adversarially robust training schemes, recent verifiably robust training  is a promising direction, which can fundamentally improve the verifiable robustness and defend against stronger attackers~\cite{wong2018scaling,mirman2018differentiable,wang2018mixtrain,gowal2018effectiveness}.

\section{Related work}
\label{sec:related}

Many defenses against adversarial examples have been proposed~\cite{gu2014towards,papernot2015distillation,cisse2017parseval,papernot2017extending,papernot2018deep,athalye2018obfuscated,buckman2018thermometer,song2017pixeldefend,xie2017mitigating,zantedeschi2017efficient}, which are followed by a sequence of stronger attacks breaking them in quick succession~\cite{papernot2016cleverhans,carlini2017towards,elsayed2018adversarial,carlini2017adversarial,moosavi2016deepfool,biggio2013evasion,ma2018characterizing,papernot2017practical,guo2017countering,he2017adversarial,athalye2017synthesizing,carlini2018efficient,pei2017deepxplore,tian2018deeptest}. We refer the interested readers to the survey by Athalye et al.~\cite{athalye2018obfuscated} for more details. In spite of large amounts of attacks, the adversarially trained model~\cite{madry2017towards} remains robust against known attacks. 
One attack paper~\cite{he2018decision} that breaks the region-based defense~\cite{cao2017mitigating} also uses the region information. However, their surrounding area estimation, unlike us, is based on sampling and thus may miss violations compared to the interval attack.

\section{Conclusion}
\label{sec:conclusion}

We propose a novel type of gradient-based attacks which is the first generic attack framework to combine sound bound propagation methods with existing gradient-based attacks. By plotting the loss distributions found by interval attack with $10^5$ random starting points, we are able to show that PGD attack is not the ultimate first-order adversary. On three adversarially trained MNIST networks, interval attack can provide on average 47\% relative improvement over PGD attacks. In MadryLab MNIST challenge~\cite{mnistchallenge}, it achieves the best performance so far. The empirical results show that there is valuable research space seeking for stronger attacks by applying tighter sound bound propagation methods and stronger first-order attacks.

\section{Acknowledgements}
This work is sponsored in part by NSF grants CNS-16-17670, CNS-18-42456, CNS-18-01426; ONR grant N00014-17-1-2010; an ARL Young Investigator (YIP) award; and a Google Faculty Fellowship. Any opinions, findings, conclusions, or recommendations expressed herein are those of the authors, and do not necessarily reflect those of the US Government, ONR, ARL, NSF, or Google.



\bibliography{ref}
\bibliographystyle{icml2019}




\end{document}